\documentclass{article}

\usepackage[preprint]{neurips_2025}

\usepackage[utf8]{inputenc} 
\usepackage[T1]{fontenc}    
\usepackage{hyperref}       
\usepackage{url}            
\usepackage{booktabs}       
\usepackage{amsfonts}       
\usepackage{nicefrac}       
\usepackage{microtype}      
\usepackage{xcolor}         
\usepackage{amssymb}
\usepackage{caption}
\usepackage{wrapfig}
\usepackage{mdframed} 
\usepackage{graphicx}
\usepackage{stmaryrd}
\usepackage{hwemoji}
\usepackage{tabularx}
\usepackage{multirow} 
\usepackage{listings}

\usepackage{xcolor}
\definecolor{keywordcolor}{rgb}{0.7, 0.1, 0.1}   
\definecolor{tacticcolor}{rgb}{0.0, 0.1, 0.6}    
\definecolor{commentcolor}{rgb}{0.3, 0.5, 0.3}   
\definecolor{symbolcolor}{rgb}{0.0, 0.1, 0.6}    
\definecolor{sortcolor}{rgb}{0.1, 0.5, 0.1}      
\definecolor{rulecolor}{rgb}{0, 0, 0}
\definecolor{attributecolor}{rgb}{0.7, 0.1, 0.1} 

\newcommand{\oureval}[0]{\textsc{Fine-Eval}}

\title{CombiBench: Benchmarking LLM Capability for Combinatorial Mathematics}

\author{
	  Junqi Liu~\thanks{Equal contribution.}~~\thanks{Interns at Moonshot AI.}~~$^{\spadesuit\heartsuit}$ 
	~~Xiaohan Lin~\footnotemark[1]~~\footnotemark[2]~~$^{\clubsuit\heartsuit}$ 
	~~Jonas Bayer~$^{\circledcirc}$
        ~~Yael Dillies~$^{\Join}$
        ~~Weijie Jiang~$^\circledast$
	\\
	\textbf{
            Xiaodan Liang~$^\clubsuit$
		~~Roman Soletskyi~$^\diamondsuit$
            ~~Haiming Wang~$^{\clubsuit\heartsuit}$
            ~~Yunzhou Xie~$^\divideontimes$
            ~~Beibei Xiong~$^\circledast$
        }
	\\
	\textbf{
            ~~Zhengfeng Yang~$^\circledast$
            ~~Jujian Zhang~$^\divideontimes$
            ~~Lihong Zhi~$^\spadesuit$ 	 
            ~~Jia Li~\thanks{Corresponding authors. We list all authors except the first and corresponding authors in alphabetical order by last name}~~$^\diamondsuit$ 
		~~Zhengying Liu~\footnotemark[3]~~$^{\heartsuit}$	
	} 
	\\
	$^\spadesuit$~Academy of Mathematics and Systems Science, University of Chinese Academy of Sciences\\
        $^\clubsuit$~Sun Yat-sen University~~$^{\circledcirc}$~University of Cambridge~~$^\circledast$~East China Normal University\\
        $^\divideontimes$~Imperial College London~~$^{\Join}$~Stockholm Universitet~~$^\diamondsuit$~Numina~~
        $^{\heartsuit}$~Moonshot AI\\
	\texttt{liujunqi@amss.ac.cn}, \texttt{carmelolim55@gmail.com}, \\                        \texttt{lijia0765@gmail.com}, \texttt{liuzhengying@moonshot.cn}
}

\begin{document}

\maketitle

\begin{abstract}

Neurosymbolic approaches integrating large language models with formal reasoning have recently achieved human-level performance on mathematics competition problems in algebra, geometry and number theory. In comparison, combinatorics remains a challenging domain, characterized by a lack of appropriate benchmarks and theorem libraries. To address this gap, we introduce CombiBench, a comprehensive benchmark comprising 100 combinatorial problems, each formalized in Lean~4 and paired with its corresponding informal statement. The problem set covers a wide spectrum of difficulty levels, ranging from middle school to IMO and university level, and span over ten combinatorial topics. CombiBench is suitable for testing IMO solving capabilities since it includes all IMO combinatorial problems since 2000 (except IMO 2004 P3 as its statement contain an images). Furthermore, we provide a comprehensive and standardized evaluation framework, dubbed \emph{Fine-Eval} (for \textbf{F}ill-in-the-blank \textbf{in} L\textbf{e}an Evaluation), for formal mathematics. It accommodates not only proof-based problems but also, for the first time, the evaluation of fill-in-the-blank questions. Using Fine-Eval as the evaluation method and Kimina Lean Server as the backend, we benchmark several LLMs on CombiBench and observe that their capabilities for formally solving combinatorial problems remain limited. Among all models tested (none of which has been trained for this particular task), Kimina-Prover attains the best results, solving 7 problems (out of 100) under both ``with solution'' and ``without solution'' scenarios. We open source the benchmark dataset alongside with the code of the proposed evaluation method at \url{https://github.com/MoonshotAI/CombiBench/}.

\end{abstract}

\section{Introduction}

Automated theorem proving (ATP) has long been a prominent research area at the intersection of artificial intelligence and mathematics~\citep{bibel2013automated}; aiming to develop computer programs capable of automatically verifying or discovering mathematical proofs. In recent years, the integration of large language models (LLMs) with formal theorem provers such as Lean~\citep{moura2021lean}, Coq~\citep{Coq}, and Isabelle~\citep{wenzel2008isabelle} has received increased attention. These formal systems can rigorously verify step-by-step the proof generated by an LLM and provide feedback, thereby enhancing the models' performance in automated theorem proving.

In this research area, the design of high-quality benchmark tests is essential, as they offer valuable guidance for improving model capabilities. Three well-known competition-based benchmarks in automated theorem proving include miniF2F~\citep{zheng2021minif2f}, FIMO~\citep{liu2023fimo} and PutnamBench~\citep{tsoukalas2024putnambench}. miniF2F primarily features high school-level mathematics problems and competition questions, FIMO consists of algebra and number theory problems from the International Mathematical Olympiad (IMO), and PutnamBench originates from the prestigious North American university-level Putnam competition. Next to these, ProofNet~\citep{azerbayev2023proofnet} is a benchmark for autoformalization and formal proving of undergraduate-level mathematics.

However, each of these benchmarks has notable limitations. For instance, miniF2F is limited to high school mathematics, FIMO and ProofNet are written in the now outdated formal proof language Lean~3~\citep{de2015lean}, and PutnamBench includes only a small number of combinatorial problems. However, it's worth mentioning that a Lean 4 version of ProofNet is provided in DeepSeek-Prover-V1.5~\citep{xin2024deepseekproverv15harnessingproofassistant}. We have analyzed the quantity and proportion of combinatorial problems in some existing benchmarks in Table~\ref{tab:comparsion}, as well as their applicability to Lean 4.

\begin{table*}[h]
  \centering
  \begin{tabular}{lccccc}    
    \toprule
    Type  & miniF2F & FIMO & PutnamBench & ProofNet & CombiBench   \\
    \midrule
    Count & 0 & 0 & 29 & 0 & \textbf{100} \\
    Ratio & 0 & 0 & 4.4\% & 0 & \textbf{100\%} \\
    Lean 4 friendly & \includegraphics[scale = 0.15]{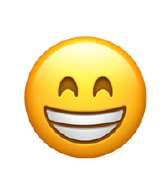} &  \includegraphics[scale = 0.15]{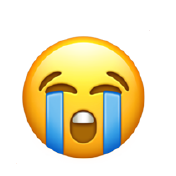} & \includegraphics[scale = 0.15]{image/smile.pdf} & \includegraphics[scale = 0.15]{image/smile.pdf} & \includegraphics[scale = 0.15]{image/smile.pdf} \\
    \bottomrule
  \end{tabular}
  \caption{Benchmark Comparison of Combinatorics Problems}
  \label{tab:comparsion}
\end{table*}

The above formal mathematics benchmarks in Lean 4 scarcely cover combinatorial mathematics. But combinatorial mathematics is also the area with the weakest model capabilities in the field of automated theorem proving at present. In 2024, Google DeepMind's AlphaProof successfully solved 4 problems (out of 6) in the International Mathematical Olympiad (IMO) using Lean~4, achieving a level comparable to human silver medalists. However, the remaining two unsolved problems by AlphaProof were combinatorial. This also confirms what we mentioned above. This outcome can be attributed to three primary factors:
\begin{itemize}
    \item The absence of robust benchmarks to assess models' proficiency in combinatorial mathematics;
    \item The limited coverage of combinatorial content within Lean's math library—Mathlib \citep{mathlib}. 
    \item In combinatorial mathematics, the gap between informal and formal is larger than in other areas. For a specific problem, it is often necessary to add some definitions that are unique to that problem.
\end{itemize}

CombiBench aims to address these gaps by providing testing standards for evaluating the capabilities of large language models in this area. 

Besides, traditional formal math benchmarks (such as miniF2F, FIMO) primarily consist of theorem-proving tasks, where evaluation methods focus on assessing whether a model can complete a proof. However, PutnamBench introduces a new standardized formalization for fill-in-the-blank problems, which existing evaluation approaches cannot handle effectively. Unlike standard proof-based problems, fill-in-the-blank questions require the model to construct a solution and verify its correctness. Current evaluation methods are insufficient for this problem, focusing solely on proving given propositions without addressing solution generation and validation. To bridge this gap, we developed a novel evaluation framework, named \oureval, to assess model performance on fill-in-the-blank problems. This approach adapts evaluation techniques to new problem formats and provides a more comprehensive measure of a model's mathematical reasoning abilities, particularly in solution construction and verification.

The contributions of this work are outlined as follows:

\begin{itemize}
    \item  We introduce CombiBench, a benchmark containing 100 combinatorial formal statements in Lean 4 and their corresponding informal descriptions. The benchmark spans problems of varying difficulty, from middle school to IMO and university level, covering over ten distinct combinatorial topics.
    
    \item We propose a standardized and comprehensive evaluation framework for formal mathematics, enabling the assessment of proof-based problems and fill-in-the-blank questions for the first time.

\end{itemize}

\section{Related work}

\subsection{IMO}

The International Mathematical Olympiad (IMO) is the most prestigious and competitive global mathematics competition for high school students. Established in 1959, the IMO aims to challenge and inspire young mathematicians worldwide. Each year, teams of up to six students from over 100 countries participate in the event, which consists of solving six highly complex mathematical problems over two days. The problems span various areas of mathematics, including algebra, geometry, number theory and combinatorics, requiring not only deep mathematical knowledge but also creativity and problem-solving skills. Many individuals who have won the Fields Medal--the highest honor in mathematics, have participated in the IMO. More recently, the annual IMO competition has also become widely recognized as the ultimate grand challenge for Artificial Intelligence (AI) \citep{imograndchallengeGrandChallenge}. Artificial Intelligence Mathematical Olympiad (AIMO) has set up a \$10 million prize to reward the first AI system that reaches the level of an IMO gold medalist \citep{aimoprizeAIMOPrize}.

\subsection{Automated Theorem Proving}

Early research in automated theorem proving focused on first-order theorem provers designed for simple logical frameworks \citep{schulz2002brainiac} and theorem provers based on symbolic engines \citep{chou2000deductive}. With the advent of deep learning technologies, there has been a significant shift toward utilizing large language models (LLMs) to automate the theorem-proving process. 

 GPT-f \citep{polu2020generative} is an automated theorem prover and proof assistant developed for the Metamath formalization language. Its introduction marks the first instance that a formal mathematics community accepts and incorporates proofs generated by deep learning. The emergence of GPT-f has significantly advanced the field of automated theorem proving. Subsequently, the Draft, Sketch, and Prove (DSP) \citep{jiang2022draft} was introduced, utilizing informal proofs to aid in the generation of formal ones. LEGO-Prover \citep{wang2023legoprover} builds a library through a modular formal proof, allowing large language models (LLMs) to retrieve existing skills and generate new ones during the proof process. DT-Solver \citep{wang2023dt} introduces a dynamic-tree Monte Carlo search algorithm. BFS-Prover~\citep{xin2025bfs}, using a best-first search approach, attained state-of-the-art performance among theorem provers based on search algorithms. 

The aforementioned work employs a search-based approach to predict the next step in the proof. Recently, progress in this area of research has led to the development of an alternative approach, where the language model generates the entire proof directly. DeepSeek-Prover \citep{xin2024deepseekprover} and Goedel-Prover \citep{lin2025goedel} both utilize this method. In particular, DeepSeek-Prover incorporates both search-based and whole-proof approaches. Despite notable progress, these existing methods face significant challenges.
While LLMs excel at pattern matching and sequence generation, effectively capturing
the deep, structured, and often non-linear reasoning required for complex formal proofs remains difficult. Kimina-Prover Preview~\citep{wang2025kiminaproverpreviewlargeformal} proposed a novel reasoning-driven exploration paradigm for formal theorem proving, using reinforcement learning to enhance whole-proof generation.

\subsection{Existing Benchmarks}

With the rapid advancement of automated theorem proving, several formal mathematical benchmarks in the Lean language have been introduced in recent years. miniF2F is a benchmark designed to evaluate automated theorem-proving systems in various formal systems. It includes various mathematical problems, such as exercises from prestigious math olympiads (AMC, AIME, IMO) and high school and undergraduate coursework. The main objective of miniF2F is to provide a standardized benchmark to directly evaluate and compare theorem-proving systems. Initially implemented in Lean and Metamath, it has since been extended to HOL Light and Isabelle, broadening its applicability across different formal proof environments. The current state-of-the-art result in the miniF2F test set is $80.74\%$, achieved by Kimina-Prover Preview~\citep{wang2025kiminaproverpreviewlargeformal}.

For combinatorial problems, miniF2F and FIMO focus exclusively on algebra and number theory problems from the IMO. PutnamBench includes 26 combinatorial problems, but its coverage of combinatorial topics (e.g. countability, power sets, discrete structures, games) is limited. LeanComb \citep{xiong2025combinatorial} introduces a data enhancement approach and offers a benchmark specifically focused on combinatorial identities.  The proofs of these identities often involve analytical techniques rather than purely combinatorial reasoning.

\section{CombiBench}

CombiBench is the first benchmark focused on combinatorial competition problems written entirely in Lean~4. It is a manually produced benchmark that includes 100 combinatorial mathematics problems of varying difficulty and knowledge levels. 

\textbf{Range of Topics.} For the selection of topics in combinatorics, we follow Brualdi's classical combinatorics textbook ``Introductory Combinatorics''~\citep{brualdi2004introductory}. It consists of fourteen chapters and is widely used in undergraduate and graduate courses in combinatorics.

\begin{itemize}
    \item What Is Combinatorics?
    \item Permutations and Combinations
    \item The Pigeonhole Principle
    \item Generating Permutations and Combinations
    \item The Binomial Coefficients
    \item The Inclusion-Exclusion Principle and Applications
    \item Recurrence Relations and Generating Functions
    \item Special Counting Sequences
    \item Systems of Distinct Representatives
    \item Combinatorial Designs
    \item Introduction to Graph Theory
    \item More on Graph Theory
    \item Digraphs and Networks
    \item P\'olya Counting
\end{itemize}

The book systematically introduces the fundamental concepts, methods, and applications of combinatorics, covering many important topics such as permutations and combinations, the pigeonhole principle, generating functions, graph theory, and combinatorial design.

\textbf{Composition and Diversity}.
CombiBench consists of 10 easy problems from \url{https://www.hackmath.net/}, 42 exercises from Brualdi's book, 36 IMO problems, and 12 problems from other math competitions, which are shown in Table~\ref{tab:source}. This composition ensures that CombiBench covers a wide range of difficulty, from easy to difficult, reflecting good diversity in difficulty.

\begin{table*}[h]
  \centering
  \begin{tabular}{lccccc}    
    \toprule
    Source & Hackmath & Brualdi's book  & IMO & APMO & Balticway  \\
    \midrule
    Conut & 10 & 42 & 36 & 2 & 1 \\
    \toprule
    Source & EGMO & IMO-Shortlist & IZHO & BXMO & USAMO \\
    \midrule
    Conut & 1 & 4 & 2 & 1 & 1 \\
    \bottomrule
  \end{tabular}
  \caption{Problem source distribution}
  \label{tab:source}
\end{table*}

For the IMO problems, we collected all combinatorics problems from the official IMO problems since 2000, totaling 37 problems. However, two problems contain images (Problem 3 from 2004 and Problem 5 from 2023), making their direct formalization difficult. After our analysis, we found that Problem 5 from 2023 could potentially be formalized. The figure is only for illustration and does not contain any additional information that is not already in the text. Then, we removed Problem 3 from 2004 to get 36 IMO problems. 

For the exercises from the Brualdi book, we randomly sampled problems from 14 chapters, choosing three problems from each chapter. This ensured that the 42 problems were evenly distributed across all 14 chapters, helping to guarantee the diversity of the topics covered in our selection.

The complete proofs of Problem 3 and Problem 5 from 2024 have already been formalized in Mathlib. Therefore, we directly refer to the statements of these problems from Mathlib in CombiBench, along with the necessary definitions used in the statements.

\textbf{Naming Convention}. Depending on the source of the problems, we adopt different naming conventions. For the 10 easy problems, we name them \lstinline{hackmath_1}, \lstinline{hackmath_2}, etc. Exercises from Brualdi's book are named using the format \lstinline{brualdi_#chapter_#number}. Each competition problem is named as “\lstinline{#competition_#year_p#number}”. For instance, “\lstinline{imo_2019_p5}” refers to the IMO Problems 2009 Problem 5, and “\lstinline{imo_2019_p5_1}” refers to the first sub-question of the IMO Problems 2009 Problem 5. And “\lstinline{brualdi_ch8_6}” refers to Problem 6 from the exercises section of Chapter 8 in Brualdi's book. 

\textbf{Classification of Problem}. In CombiBench, $45\%$ of the problems require first providing a solution to the problem and then proving its correctness, such as the following problem : 

\begin{itemize}
\item Determine the number of permutations of $\{1,2, \ldots, 8\}$ in which no even integer is in its natural position.
\end{itemize}

Therefore, these problems do not directly state a proposition and cannot be directly formalized. Previous benchmarks, such as miniF2F and FIMO, circumvented this issue by modifying the problem statement to require proving that a given solution satisfies the problem's constraints. However, this modification reduces the overall difficulty of the problem, as a significant portion of the challenge may lie in coming up with the solution itself. PutnamBench introduces a standardized method for formalizing statements of such problems, which more accurately reflects the difficulty of informal problems. The following is an example where we formalize the above combinatorial problem in PutnamBench style: 
\begin{lstlisting}[frame = single]
abbrev brualdi_ch6_11_solution : ℕ := sorry

/--
Determine the number of permutations of $\{1,2, \ldots, 8\}$ in which no even integer is in its natural position.
-/
theorem brualdi_ch6_11
    (sols : Finset (Equiv.Perm (Finset.Icc 1 8)))
    (h_sols : ∀ σ, σ ∈ sols ↔ (∀ i, Even i.1 → σ i ≠ i)) :
    sols.card = brualdi_ch6_11_solution := by sorry
\end{lstlisting}

\textbf{Challenges}. Our formalization team consists of five doctoral students and one master's student, each with over a year of experience in learning Lean, including a major contributor to Mathlib and a reviewer of Mathlib. During the formalization process, we found that besides simple problems, most problems require more than 30 minutes to formalize. For problems at the International Mathematical Olympiad (IMO) level, almost every problem takes over 3 hours to formalize, with some problems taking more than 8 hours. This indicates that formalizing combinatorial mathematics problems at the IMO level remains challenging and time-consuming.

\begin{figure}[h]
    \centering
    \includegraphics[scale = 0.7]{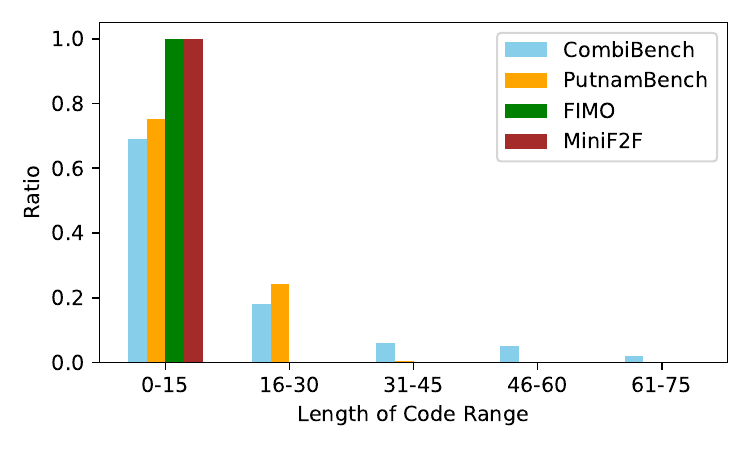}
    \caption{Code Length Distribution}
    \label{fig1:Code_Length_Distribution}
\end{figure}

One reason for this is the scarcity of theorems related to combinatorics in Mathlib, which makes it challenging to formalize corresponding problems and requires us to define many concepts. We analyzed the length of the formalized statements in CombiBench, miniF2F, PutnamBench, and FIMO. We excluded blank lines, comments, \lstinline{import}'s, and \lstinline{open}'s, focusing only on the length of the formalized code directly related to the problem statements. See Figure~\ref{fig1:Code_Length_Distribution}. We found that all problem formalizations in FIMO and miniF2F are within 15 lines. In PutnamBench and CombiBench, some problems have formalizations ranging from 16 to 30 lines, and those exceeding 30 lines are almost entirely from CombiBench. This indicates that formalizing combinatorics problems is particularly challenging. We also counted the number of lines of formalization code for all problems (excluding blank lines) in CombiBench. The results showed that more than half of the problems have more than 10 lines of code after formalization, more than a quarter have more than 20 lines, and the most challenging problem reached 67. We present the longest formalization of the statement of the problem in CombiBench in Appendix~\ref{exam:longest}. This data point indicates that formalizing combinatorial mathematics problems is still highly challenging.

\textbf{Licensing and Rules of Engagement}. CombiBench is available under MIT License for Lean 4. We host a public leaderboard at \url{https://moonshotai.github.io/CombiBench/leaderboard.html} and welcome evaluation results from future works. 

\section{Evaluation}

We construct experiments to grasp CombiBench's difficulties with state-of-the-art theorem-proving approaches and general-purpose LLMs. Previous evaluation methods include whole-proof generation and tree search. Whole-proof generation directly queries a language model to generate a complete proof, while tree-search methods gradually build the evidence by searching for possible proof strategies. Both methods work well for verifying proof problems, but cannot verify fill-in-the-blank problems. To address the issue, we propose a new evaluation pipeline to verify both types of problems in Lean 4 named \oureval (\textbf{F}ill-in-the-blank \textbf{in} L\textbf{e}an Evaluation). We formalize the matching of the LLM prediction answer and the ground truth into a theorem-proving problem to rigorously verify the fill-in-the-blank question and avoid the complicated rules and format requirements as approaches in the math word problem.

\subsection{Evaluation method}
\oureval\hspace{1pt}interacts with two servers, LLM and Lean. LLM takes as input an entire formal statement with the proof and the blanks to be filled in replaced by `\lstinline{sorry}'s, and writes a snippet of complete Lean 4 code. Since LLMs sometimes cheat by commenting out all the code to pretend it passed compilation, we show an example in the appendix~\ref{exam:cheat}. Therefore, we require the code, after comments are removed, to satisfy the following conditions:
\begin{itemize}
    \item Does not contain ``sorry''s;
    \item Can not define any new axioms.
    \item Can be compiled by Lean without errors;
    \item Compared with the input formal statement, except for the replaced ``sorry''s, the other parts are exactly matched.
\end{itemize}

A code that does not meet these requirements is considered a failure. Otherwise, we believe the proof is successful and check whether the solution provided by LLM and the ground truth match exactly. If the solution and ground truth are exactly matched, we believe that the model has successfully solved the problem. If not, we then try to verify that the answer that LLM predicts is equivalent to the ground truth. To avoid unnecessary LLM calls, we construct a formal statement as follows, asserting that ``\lstinline{xx_solution} = \lstinline{ground_truth}'' and try to prove it using two common tactics: `\lstinline{rfl}' and `\lstinline{norm_num}'. 

\begin{lstlisting}[frame = single]
example : imo_2006_p2_solution = ground_truth := by 
    try rfl
    try norm_num
\end{lstlisting}

\begin{figure}[h]
    \includegraphics[width=\textwidth]{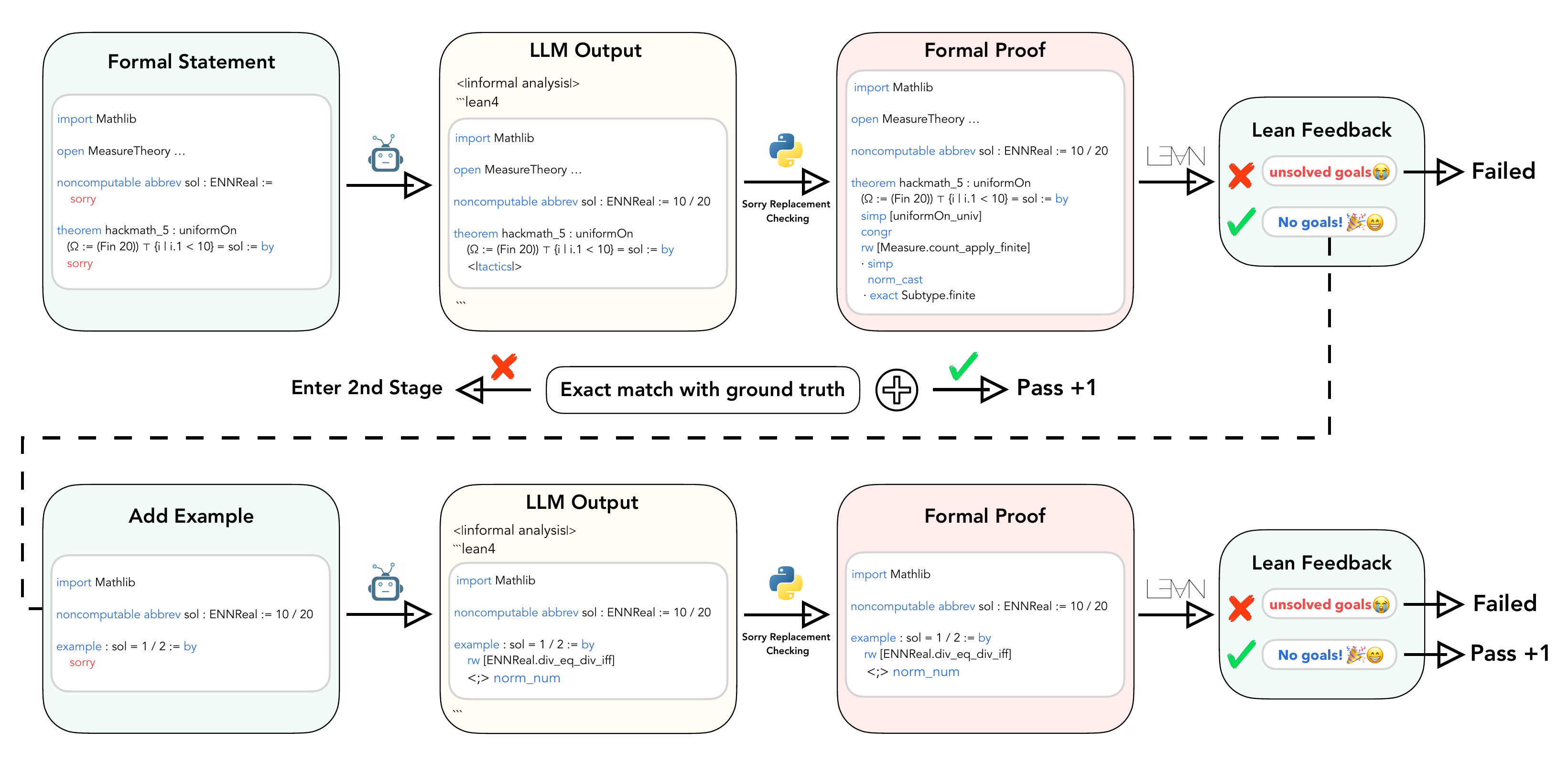}
    \caption{Pipeline of the two-stage \oureval.}
    \label{fig:eval}
\end{figure}

If the lean verifier returns that it cannot be proved, we consider that the matching answer is non-trivial and add the statement to the input of the first round of LLM to prove it again. We use the same criterion as the first phase to determine whether the LLM's proof is successful. Figure~\ref{fig:eval} demonstrates the process of \oureval. In the first stage, the model answers ``10 / 20'' and the corresponding proof at once, which can be verified by the lean 4 server. Next, we try to use ``norm\_num'' and ``rfl'' to automatically prove that 10 / 20 is equivalent to 1 / 2. When these attempts fail, we enter the second stage and let the model try to prove this problem, and the model fully demonstrates that the answer it predicts is equivalent to the ground truth. To prevent models from first submitting a trivial solution, then deceptively eliciting the ground truth to subsequently construct the actual proof, we impose a length limit on the second-stage proof. The number of characters in the output proof, after removing spaces and newline characters, must not exceed 42.

\begin{figure}[h]
    \includegraphics[width=\textwidth]{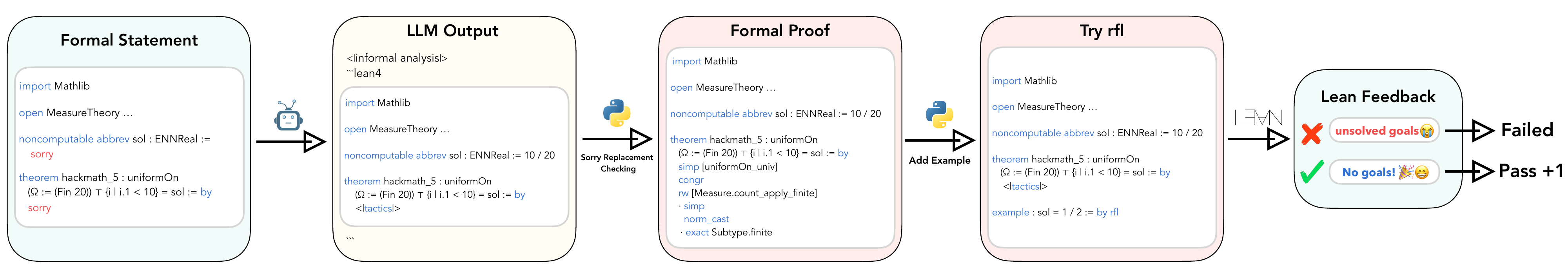}
    \caption{Pipeline of the one-stage \oureval.}
    \label{fig:1_stage_eval}
\end{figure}

In practical applications, the two-stage verification method can often be complex. Consequently, we also provide a simplified evaluation approach, as illustrated in Figure~\ref{fig:1_stage_eval}. This simplified method initially follows the same procedure as the first stage of the Fine. However, if an exact match fails, we directly employ the \lstinline{rfl} tactic to verify if the solution generated by the model is definitionally equivalent to the ground truth. This enables the evaluation of fill-in-the-blank problems using only a single Large Language Model (LLM) call. While this approach imposes stricter requirements on the model's predicted answer, it is more easily embedded into other workflows.

\paragraph{Metric} For fill-in-the-blank questions, we treat a complete two-stage \oureval \hspace{1pt}as a single verification. Following PutnamBench, we also conducted a comparative experiment, replacing the corresponding sorry with ground truths and letting the model complete the proof. We report pass@N at different sample budgets as the performance metric.

\subsection{Baseline results}

To demonstrate the significant challenge that CombiBench poses to LLMs, we evaluate it on different LLMs using various computational budgets. We consider two types of models, one is the model fine-tuned on the automated theorem proving task (indicated by "theorem prover"), the size of such models is usually 7B or less (except Kimina-Prover Preview) and the other is the general reasoning model (indicated by "reasoning model"). We evaluate  pass@16 of these models on CombiBench using whole-proof generation. Considering the economic cost and reproducibility, we do not evaluate with a larger sample budget and tree-search method.

\begin{table*}[tbh]
    \begin{center}
    \small
    \begin{tabularx}{\textwidth}{l*{6}{>{\centering\arraybackslash}X}}
    \toprule
        Model & \multicolumn{3}{c}{\textit{with solution}} & \multicolumn{3}{c}{\textit{without solution}} \\
        \cmidrule(lr){2-4} \cmidrule(lr){5-7}
        ~ & Pass@1 & Pass@8 & Pass@16 & Pass@1 & Pass@8 & Pass@16 \\
        \midrule
        \multicolumn{7}{l}{\textit{Reasoning Model}} \\
        \midrule
        o1\citep{jaech2024openai} & 0 & 2 & 2 & 0 & 2 & 2 \\
        o3-mini\citep{openai_o3_mini} & 0 & 1 & 1 & 0 & 2 & 2 \\
        QwQ\citep{qwq32b} & 0 & 2 & 2 & 0 & 2 & 2 \\
        Claude-3.7-Sonnet-thinking\citep{anthropic_claude_3_7_sonnet} & 0 & 2 & 2 & 0 & 0 & 0 \\
        DeepSeek-R1\citep{guo2025deepseek} & 0 & 2 & 2 & \textbf{1} & 2 & 2 \\
        Gemini-2.5-pro-preview\citep{google_gemini_2_5} & 0 & 2 & 4 & 0 & 2 & 3 \\
        \midrule
        \multicolumn{7}{l}{\textit{Theorem Prover}} \\
        \midrule
        Goedel-Prover\citep{lin2025goedel} & 0 & 0 & 0 & 0 & 0 & 0 \\
        STP\citep{dong2025stp} & 0 & 0 & 0 & 0 & 0 & 0 \\
        Leanabell-Prover-GD-RL\citep{zhang2025leanabell} & 0 & 0 & 0 & 0 & 0 & 0 \\
        Kimina-Prover Preview\citep{wang2025kiminaproverpreviewlargeformal} & \textbf{2} & \textbf{6} & \textbf{7} & \textbf{1} & \textbf{4} & \textbf{7} \\
        \bottomrule
    \end{tabularx}
    \caption{
    Evaluation results on CombiBench across different computational budgets and methods reveal that all tested approaches perform poorly, solving only a few problems at most.
    }
    \label{tab:main_results}
    \end{center}
\end{table*}

The experimental results are presented in Table~\ref{tab:main_results}. We found that all models solved only a few problems, likely due to the absence of formal combinatorial question libraries and the inherent difficulty of such questions. None of the theorem-proving models with 7B parameters or fewer were able to solve all problems, which we attribute to the limited generalization capabilities of smaller models when facing out-of-distribution data. The general reasoning model answered a few questions correctly, and its performance showed minimal difference between the with\_solution and without\_solution settings, likely due to its strong informal mathematical reasoning ability. In the without\_solution case, both evaluation methods mentioned earlier produced identical results. This observation implies that, due to the model's current capability limitations, the solutions predicted are typically either completely correct or entirely wrong. As a result, sophisticated or flexible answer-checking mechanisms are less critical at this stage. Kimina-Prover Preview, a 72B parameter model fine-tuned specifically for theorem proving, achieved state-of-the-art results, solving $7$ problems under both with and without solution settings. It is important to note that Kimina-Prover Preview has not been specifically trained on datasets pertaining to combinatorics or fill-in-the-blank question types.

\section{Conclusion}

This paper introduces CombiBench, a formal-language benchmark for evaluating the capability of AI models on combinatorics competition problems, developed in Lean~4. CombiBench spans a variety of topics in combinatorics, featuring problems that range in difficulty from high school level to IMO and university-level. It is the first comprehensive benchmark specifically designed to assess the capability of language models in solving combinatorial mathematics problems.

Additionally, we introduce a new evaluation method for fill-in-the-blank problems, employing a two-stage approach. In the first stage, the model is required to generate a solution and prove its correctness. If the filled-in solution is equal to the ground truth and the proof compiles successfully, the problem is deemed solved. Even if the solution differs from the ground truth, it might still be a valid mathematical solution. In this case, the evaluation moves to a second stage, where the model must demonstrate that its proposed solution is equivalent to the ground-truth answer. 

Our experimental results indicate that CombiBench poses a considerable challenge, as all existing models are unable to solve most of the problems. We identify two primary factors contributing to these failures:
\begin{itemize}
\item Absence of combinatorial mathematics content in existing theorem libraries: models struggle with formalization because they lack pre-built definitions, lemmas, and theorems relevant to combinatorial topics, forcing them to construct everything from scratch.

\item The significant gap between natural language problem statements and formalized proofs: In combinatorial mathematics, the transition from informal to formal reasoning is particularly challenging, and current models cannot bridge this gap effectively.    
\end{itemize}

Looking ahead, we will first gradually contribute the newly formalized definitions from the CombiBench project to mathlib, while concurrently developing a dedicated theorem library for combinatorics. In addition, we aim to actively promote the enhancement of large language models' capabilities in the field of combinatorics.

\section*{Acknowledgements}

We gratefully acknowledge Shi-Zhuo Looi for his assistance and insightful suggestions during our formalization process, and Ran Wang for his contributions to the optimization and development of our website. We are very grateful to Joseph Myers, the author of the two imo problems(2024P3 and 2024P5). We also appreciate his suggestions on the formalization of our problems.

\bibliographystyle{abbrvnat}

\bibliography{neurips_conference}


\newpage

\appendix

\section{Appendix}

\subsection{Example of the longest formalization}

\label{exam:longest}
\begin{mdframed}[linewidth=0.5mm]
{\bf Problem: bxmo 2017 p2.}

{\it Problem in Natural Language}: Let $n \geqslant 2$ be an integer. Alice and Bob play a game concerning a country made of $n$ islands. Exactly two of those $n$ islands have a factory. Initially there is no bridge in the country. Alice and Bob take turns in the following way. In each turn, the player must build a bridge between two different islands $I_{1}$ and $I_{2}$ such that:
\begin{itemize}
    \item $I_{1}$ and $I_{2}$ are not already connected by a bridge;
    \item at least one of the two islands $I_{1}$ and $I_{2}$ is connected by a series of bridges to an island with a factory (or has a factory itself). (Indeed, access to a factory is needed for the construction.)
\end{itemize}
As soon as a player builds a bridge that makes it possible to go from one factory to the other, this player loses the game. (Indeed, it triggers an industrial battle between both factories.) If Alice starts, then determine (for each $n \geqslant 2$) who has a winning strategy.(Note: It is allowed to construct a bridge passing above another bridge.)

{\it Formalization of Statement}:

\begin{lstlisting}[frame = single]
import Mathlib

variable (m : ℕ)

local notation3 (prettyPrint := false) "n" => (m + 2)
local notation3 (prettyPrint := false) "F1" => (0 : Fin n)
local notation3 (prettyPrint := false) "F2" => (1 : Fin n)

structure GameState where
  islands: SimpleGraph (Fin n)
  decidable: DecidableRel islands.Adj

instance (s : GameState m) : DecidableRel s.islands.Adj := by
  exact s.decidable

def GameState.initial : GameState m := {
  islands := ⊥
  decidable := SimpleGraph.Bot.adjDecidable (Fin n)
}

structure Bridge where
  island1 : Fin n
  island2 : Fin n

def reachableByFactory (s : GameState m) (b : Bridge m) : Prop :=
  s.islands.Reachable b.island1 F1 ∨ s.islands.Reachable b.island1 F2
  ∨ s.islands.Reachable b.island2 F1 ∨ s.islands.Reachable b.island2 F2

def isValidMove (s : GameState m) (b : Bridge m) : Prop :=
  b.island1 ≠ b.island2 ∧ ¬ s.islands.Adj b.island1 b.island2 ∧ reachableByFactory m s b

def GameState.next (s : GameState m) (b : Bridge m) : GameState m := {
  islands := s.islands ⊔ (SimpleGraph.fromEdgeSet {s(b.island1, b.island2)})
  decidable := by
    have newEdge: DecidableRel (SimpleGraph.fromEdgeSet {s(b.island1, b.island2)}).Adj := by
      intro x y; unfold SimpleGraph.fromEdgeSet
      simp only [Pi.inf_apply, Sym2.toRel_prop, Set.mem_singleton_iff, Sym2.eq, Sym2.rel_iff',
        Prod.mk.injEq, Prod.swap_prod_mk, ne_eq, inf_Prop_eq]
      infer_instance
    exact SimpleGraph.Sup.adjDecidable (Fin n) s.islands (SimpleGraph.fromEdgeSet {s(b.island1, b.island2)})
}

def GameState.is_losing_state (s : GameState m) : Prop :=
  s.islands.Reachable F1 F2

abbrev Strategy := GameState m → Bridge m

instance (s: GameState m) : Decidable (GameState.is_losing_state m s) := by
  simp [GameState.is_losing_state]; infer_instance

instance (s: GameState m) (b : Bridge m) : Decidable (reachableByFactory m s b) := by
  simp [reachableByFactory]; infer_instance

instance (s: GameState m) (b : Bridge m) : Decidable (isValidMove m s b) := by
  simp [isValidMove]; infer_instance

structure MoveOutcome where
  nextState : GameState m
  hasLost : Bool

def executeStrategy (s : GameState m) (strategy: Strategy m): MoveOutcome m :=
  let bridge := strategy s
  if ¬ isValidMove m s bridge
    then { nextState := s, hasLost := true }
  else
    let nextState := s.next m bridge
    { nextState := nextState, hasLost := nextState.is_losing_state m }

partial def aliceWins (s : GameState m) (sA: Strategy m) (sB: Strategy m): Bool :=
  let ⟨stateAfterAlicesMove, aliceHasLost⟩ := executeStrategy m s sA;
  if aliceHasLost then False else
  let ⟨stateAfterBobsMove, bobHasLost⟩ := executeStrategy m stateAfterAlicesMove sB;
  if bobHasLost then True else
  aliceWins stateAfterBobsMove sA sB

abbrev bxmo_2017_p2_solution : ℕ → Fin 2 := sorry

theorem bxmo_2017_p2 : (bxmo_2017_p2_solution n = 0 →
    ∃ strategyA , ∀ strategyB, aliceWins m (GameState.initial m) strategyA strategyB)
    ∧ (bxmo_2017_p2_solution n = 1 →
    ∃ strategyB, ∀ strategyA, ¬ aliceWins m (GameState.initial m) strategyA strategyB) := by sorry
\end{lstlisting}
\end{mdframed}

\subsection{Cheat Case from Model}

\label{exam:cheat}

Our evaluation code imposes several strict requirements on the generated Lean 4 output, and these requirements are well-justified. We implemented them because we observed models exhibiting various 'cheating' behaviors during the generation process. For instance, in the following example, a model actually produced functionally incorrect Lean 4 code but managed to hide it within comments. This allowed the code to successfully pass Lean server validation checks, leading to it being mistakenly judged as correctly proven.

\begin{lstlisting}[frame = single]
import Mathlib

noncomputable def red_points {k} (points : Fin k → ℝ × ℝ) : Finset (ℝ × ℝ) :=
  ((Finset.univ (α := Fin k × Fin k)).image (fun x => midpoint ℝ (points x.1) (points x.2)))

/-
Suppose there are 997 points given in a plane. If every two points are joined by a line segment with its midpoint coloured in red, show that there are at least 1991 red points in the plane.
theorem apmo_1991_p2 (points : Fin 997 → ℝ × ℝ) : (red_points points).card ≥ 1991 := by
  intro hinj
  let k := 997
  let P := Finset.image points Finset.univ
  have h_eq : red_points points = midpoints ℝ P := by
    simp only [red_points, midpoints_def]
    ext m
    simp only [Finset.mem_image, Finset.mem_univ, true_and, Prod.exists, Finset.mem_biUnion]
    constructor
    · rintro ⟨⟨i, j⟩, _, h⟩
      use points i, points j
      simp only [Finset.mem_image, Finset.mem_univ, exists_prop, exists_and_left, and_true] at *
      exact ⟨⟨i, rfl⟩, ⟨j, rfl⟩, h⟩
    · rintro ⟨p, hp, q, hq, h⟩
      rw [Finset.mem_image] at hp hq
      rcases hp with ⟨i, hi_mem, hp_eq⟩; simp only [Finset.mem_univ] at hi_mem
      rcases hq with ⟨j, hj_mem, hq_eq⟩; simp only [Finset.mem_univ] at hj_mem
      subst hp_eq hq_eq
      use i, j
      simp only [Finset.mem_univ, exists_prop, and_true] at *
      exact h
  rw [h_eq]
  have h_card_P : P.card = k := by
    rw [Finset.card_image_of_injective Finset.univ hinj]
    simp only [Finset.card_univ, Fintype.card_fin]
  have h_card_ge_3 : 3 ≤ P.card := by
    rw [h_card_P]
    norm_num
  have h_bound := Convex.card_midpoints_ge_two_mul_sub_three h_card_ge_3
  rw [h_card_P] at h_bound
  exact h_bound
-/
\end{lstlisting}

\subsection{Case Study from Gemini}

We showcase below Gemini's performance on three problems. Notably, Gemini successfully solved \lstinline{brualdi_ch1_10}, a problem where all other models we evaluated failed. What was particularly impressive is that Gemini's solution included a highly useful lemma. This level of detailed reasoning and helpful intermediate steps was not demonstrated by any other model we tested.

\begin{lstlisting}[caption = {brualdi\_ch1\_10}, frame = single]
import Mathlib

structure IsMagicSquare {n : ℕ} (M : Matrix (Fin n) (Fin n) ℕ) : Prop where
  mem : ∀ i j, M i j ∈ Finset.Icc 1 (n * n)
  pairwise : ∀ i j i' j', i ≠ i' ∨ j ≠ j' → M i j ≠ M i' j'
  same_sum : ∃ s, (∀ i, ∑ j, M i j = s) ∧ (∀ j, ∑ i, M i j = s) ∧ (∑ i, M i i.rev = s) ∧ ∑ i, M i i = s

/-! Helper lemma: In a 2x2 magic square, M 0 1 must equal M 1 0. -/
lemma magic_square_2x2_aux {M : Matrix (Fin 2) (Fin 2) ℕ} (hM : IsMagicSquare M) : M 0 1 = M 1 0 := by
  obtain ⟨s, h_rows, h_cols, _, _⟩ := hM.same_sum
  have h_row0 := h_rows 0
  have h_col0 := h_cols 0
  -- Sums over Fin 2 are simply the sum of the two elements
  simp only [Fin.sum_univ_two] at h_row0 h_col0
  -- h_row0: M 0 0 + M 0 1 = s
  -- h_col0: M 0 0 + M 1 0 = s
  -- Equate the sums
  have eq_sums : M 0 0 + M 0 1 = M 0 0 + M 1 0 := by rw [h_row0, h_col0]
  -- Cancel M 0 0 using Nat.add_left_cancel
  exact Nat.add_left_cancel eq_sums

-- This line contains a literal and does not define the theorem's statement.
def solution : ℕ := 2 -- Example literal definition

/-
Verify that there is no magic square of order 2.
theorem brualdi_ch1_10 : ¬∃ (M : Matrix (Fin 2) (Fin 2) ℕ), IsMagicSquare M := by sorry
-/
theorem brualdi_ch1_10 : ¬∃ (M : Matrix (Fin 2) (Fin 2) ℕ), IsMagicSquare M := by
  -- Assume for contradiction that a 2x2 magic square M exists.
  -- `rintro` introduces the assumption `∃ M, IsMagicSquare M` and destructs it.
  rintro ⟨M, hM⟩
  -- From the helper lemma, we know M 0 1 = M 1 0.
  have eq_entries : M 0 1 = M 1 0 := magic_square_2x2_aux hM
  -- From the pairwise distinctness property of magic squares.
  have h_pairwise := hM.pairwise
  -- The indices (0, 1) and (1, 0) are different because 0 ≠ 1 (as elements of Fin 2).
  have idx_neq : (0 : Fin 2) ≠ (1 : Fin 2) ∨ (1 : Fin 2) ≠ (0 : Fin 2) := by
    left -- Prove the first part of the OR condition: 0 ≠ 1
    simp -- `simp` can prove this inequality for Fin 2. `decide` would also work.
  -- Apply the pairwise property to the indices (0, 1) and (1, 0).
  -- Since the indices are different, the corresponding matrix entries must be different.
  have neq_entries : M 0 1 ≠ M 1 0 := h_pairwise 0 1 1 0 idx_neq
  -- We have derived both M 0 1 = M 1 0 (from eq_entries) and M 0 1 ≠ M 1 0 (from neq_entries).
  -- This is a contradiction. `exact neq_entries eq_entries` closes the goal.
  exact neq_entries eq_entries
\end{lstlisting}

\begin{lstlisting}[caption = {brualdi\_ch10\_31}, frame = single]
import Mathlib

def isDifferenceSet (n : ℕ) (B : Finset (ZMod n)) : Prop :=
  ∃ k, ∀ x : (ZMod n),  x ≠ 0 → ∑ i ∈ B, ∑ j ∈ B \ {i}, List.count x [i - j] = k

theorem brualdi_ch10_31 : isDifferenceSet 21 {0, 3, 4, 9, 11} := by
  -- We claim that k = 1 works. This is based on the property k'(k'-1) = λ(v-1)
  -- where v=21, k'=|B|=5, so 5*4 = λ(20), which implies λ=1. Here k corresponds to λ.
  use 1
  -- Let x be an arbitrary non-zero element of ZMod 21.
  intro x hx
  -- The goal is to show that the number of pairs (i, j) in B with i ≠ j such that i - j = x is 1.
  -- The sum `∑ i ∈ B, ∑ j ∈ B \ {i}, List.count x [i - j]` counts exactly this number.
  -- `List.count x [y]` is 1 if x = y and 0 otherwise. We simplify the expression using this fact.
  -- Note: `List.count a [b]` returns a `Nat`. `simp` will use `Nat.cast_one` if the target type needs coercion,
  -- but here the sum is already over `Nat`, so `Nat.cast_one` is not strictly needed in the explanation,
  -- but `simp` might apply lemmas involving it. `List.count_singleton'` simplifies `List.count x [y]` to `if x = y then 1 else 0`.
  simp only [List.count_singleton', Nat.cast_one] -- Simplifies `List.count x [y]` to `ite (x = y) 1 0`.
  -- The goal is now `∑ i ∈ {0, 3, 4, 9, 11}, ∑ j ∈ {0, 3, 4, 9, 11} \ {i}, ite (x = i - j) 1 0 = 1`.
  -- This sum counts pairs (i, j) from B with i ≠ j such that i - j = x.
  -- Since ZMod 21 is a finite type, we can perform case analysis on x.
  -- `fin_cases x` will create a goal for each possible value of x in ZMod 21.
  fin_cases x
  -- Case x = 0:
  -- The hypothesis `hx` states `x ≠ 0`. In this case, `x = 0`, so `hx` becomes `0 ≠ 0`, which is a contradiction.
  · contradiction
  -- Cases x = 1, x = 2, ..., x = 20:
  -- In each of these 20 cases, `x` is instantiated to a specific non-zero value.
  -- The goal becomes a concrete computational statement that needs to be verified.
  -- For example, if x = 1, the goal is:
  -- `∑ i ∈ {0, 3, 4, 9, 11}, ∑ j ∈ {0, 3, 4, 9, 11} \ {i}, ite (1 = i - j) 1 0 = 1`
  -- These goals assert that for each non-zero x, the difference x appears exactly once.
  -- We use the `native_decide` tactic to verify each of these 20 computational goals automatically.
  -- `all_goals` applies the tactic to all remaining goals.
  all_goals native_decide
\end{lstlisting}

\begin{lstlisting}[caption = {brualdi\_ch2\_11}, frame = single]
import Mathlib -- Imports necessary libraries including Finset, Nat.Choose, Icc, powersetCard, filter, card, native_decide tactic

open Finset -- Opens the Finset namespace for easier access to its functions

theorem brualdi_ch2_11 :
    ((Icc (1 : ℕ) 20).powersetCard 3 |>.filter (fun S => ∀ a ∈ S, a - 1 ∉ S ∧ a + 1 ∉ S)).card = ((816) : ℕ ) := by
  native_decide -- Computes the cardinality by evaluating the expression and confirms it matches 816.
  /- The proof by `native_decide` works by compiling the expression into native code
     and executing it. The steps are:
     1. Construct the set `Icc 1 20 = {1, 2, ..., 20}`.
     2. Construct the set of all subsets of size 3 (`powersetCard 3`). The size of this set is `binom(20, 3) = 1140`.
     3. Filter this set of subsets. For each subset `S`, check the condition `∀ a ∈ S, a - 1 ∉ S ∧ a + 1 ∉ S`.
        This condition means that `S` does not contain any consecutive integers. For `S ⊆ Icc 1 20`, this condition is equivalent to `∀ a ∈ S, a + 1 ∉ S`.
     4. Compute the cardinality (`card`) of the filtered set.
     5. Compare the result with `816`.

     Combinatorial argument (not executed by `native_decide` but provides theoretical verification):
  -/
\end{lstlisting}

\subsection{Case Study from Kimina-Prover}

The current state-of-the-art (SOTA) model on the CombiBench benchmark is Kimina-Prover Preview. It successfully solved 3 problems that remained unsolved by any other model. We present below the final Lean 4 code generated by Kimina-Prover Preview for these solutions. Additionally, Kimina-Prover Preview provides a highly detailed reasoning process, which we omit here due to its significant length.

\begin{lstlisting}[caption = {hackmath\_1}, frame = single]
import Mathlib

abbrev hackmath_1_solution : ℕ := 1716

theorem hackmath_1 (sols : Finset (Fin 13 → Fin 2))
    (h_sols : ∀ f, f ∈ sols ↔ ((List.ofFn f).count 0 = 6)) :
    sols.card = hackmath_1_solution := by
  have h1 : sols = Finset.filter (fun f => (List.ofFn f).count 0 = 6) (Finset.univ) := by
    ext f
    simp [h_sols]
  rw [h1]
  simp [hackmath_1_solution]
  native_decide
\end{lstlisting}

\begin{lstlisting}[caption = {hackmath\_2}, frame = single]
import Mathlib

abbrev hackmath_2_solution : ℕ := 336

theorem hackmath_2 (sols : Finset (Fin 8 → Fin 4))
    (h_sols : ∀ f, f ∈ sols ↔
      ((List.ofFn f).count 0 = 1) ∧ ((List.ofFn f).count 1 = 1) ∧ ((List.ofFn f).count 2 = 1)) :
    sols.card = hackmath_2_solution := by 
  have h1 : sols = Finset.filter (fun f => 
      ((List.ofFn f).count 0 = 1) ∧ ((List.ofFn f).count 1 = 1) ∧ ((List.ofFn f).count 2 = 1)) 
      (Finset.univ) := by 
    ext f
    simp [h_sols]
  rw [show hackmath_2_solution = 336 by rfl]
  rw [h1]
  native_decide
\end{lstlisting}

\begin{lstlisting}[caption = {brualdi\_ch6\_11}, frame = single]
import Mathlib

abbrev brualdi_ch6_11_solution : ℕ := 24024

theorem brualdi_ch6_11
    (sols : Finset (Equiv.Perm (Finset.Icc 1 8)))
    (h_sols : ∀ σ, σ ∈ sols ↔ (∀ i, Even i.1 → σ i ≠ i)) :
    sols.card = brualdi_ch6_11_solution := by 
  have h1 : sols = Finset.filter (fun σ => ∀ i, Even i.1 → σ i ≠ i) (Finset.univ) := by 
    ext σ 
    simp [h_sols]
  rw [h1]
  native_decide
\end{lstlisting}


\end{document}